# Graphical Abstract

**SIBILA: A novel interpretable ensemble of general-purpose machine learning models applied to medical contexts**

Antonio Jesús Banegas-Luna, Horacio Pérez-Sánchez

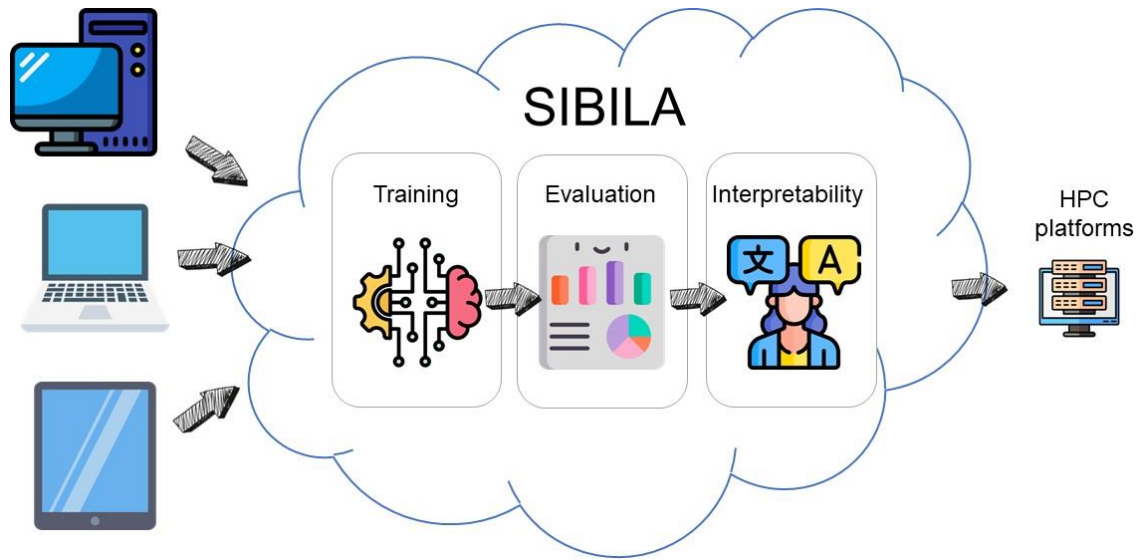

# Highlights

**SIBILA: A novel interpretable ensemble of general-purpose machine learning models applied to medical contexts**

Antonio Jesús Banegas-Luna, Horacio Pérez-Sánchez

- Machine learning is a powerful approach for the analysis of medical data.

- Black-box models are not useful in critical contexts, such as medicine. Interpretability is therefore a must.

- SIBILA has been developed as an ensemble of ML models which can be easily trained, evaluated and explained in a simple workflow.

- As interpretability is very computationally expensive, it has been accelerated by means of HPC. SIBILA can be run either locally or on HPC platforms.

- To deal with differently configured HPC infrastructures, SIBILA has been containerized with Singularity. This makes of SIBILA a very portable tool.

# SIBILA: A novel interpretable ensemble of general-purpose machine learning models applied to medical contexts

Antonio Jesús Banegas-Luna[a], Horacio Pérez-Sánchez[a]

[a]*Structural Bioinformatics and High-Performance Computing Research Group (BIO-HPC), Universidad Católica de Murcia (UCAM), Campus de los Jerónimos, Murcia, 30107, Murcia, Spain*

**Abstract**

Personalized medicine remains a major challenge for scientists. The rapid growth of Machine learning and Deep learning has made them a feasible alternative for predicting the most appropriate therapy for individual patients. However, the need to develop a custom model for every dataset, the lack of interpretation of their results and high computational requirements make many reluctant to use these methods.

Aiming to save time and bring light to the way models work internally, SIBILA has been developed. SIBILA is an ensemble of machine learning and deep learning models that applies a range of interpretability algorithms to identify the most relevant input features. Since the interpretability algorithms may not be in line with each other, a consensus stage has been implemented to estimate the global attribution of each variable to the predictions. SIBILA is containerized to be run on any high-performance computing platform. Although conceived as a command-line tool, it is also available to all users free of charge as a web server at https://bio-hpc.ucam.edu/sibila. Thus, even users with few technological skills can take advantage of it.

SIBILA has been applied to two medical case studies to show its ability to predict in classification problems. Even though it is a general-purpose tool, it has been developed with the aim of becoming a powerful decision-making tool for clinicians, but can actually be used in many other domains. Thus, other two non-medical examples are supplied as supplementary material to prove that SIBILA still works well with noise and in regression problems.

*Keywords:*

1. **Introduction**

The rapid development of technologies has helped artificial intelligence (AI) become a well-known and reliable tool for researchers in academia and industry. Its ability to analyze vast amounts of data has become a powerful tool in science and business. Looking for repetitive patterns among such datasets is a complex but necessary task that needs to be done to extract knowledge from past events. Once the rules managing raw data are identified, AI models can use them to make predictions about new unexplored samples.

Machine learning (ML) and, its subtype, Deep learning (DL) are two typical approaches of AI [1]. Both types of models are flexible enough to analyze a range of datasets, including tabular data, time series and images. This adaptability to different contexts has propelled their application into traditional and fundamental areas of science, such as biology [2, 3], chemistry [4, 5] and medicine [6]. Not only classical scientific areas can profit from ML and DL, but also new and related multidisciplinary fields. This is the case for genomics [7, 8], bioinformatics [9] and drug discovery [10, 11, 12], to name a few.

Of all the scientific areas mentioned, medicine is probably the one with the greatest visibility in society. Advances related to medicine are frequently considered highly relevant, meaning any help is always welcome. Consequently, the way ML and DL help learn how diseases could be cured or treated is a timely topic. As a result, more than a few examples of both methods applied to medicine can be found in the literature. Pérez-Gandía et al. [13] developed a decision support system that helps patients with type 1 diabetes mellitus to monitor their glucose levels daily. As such, abnormal situations can be anticipated, and measures can be taken earlier than with a traditional procedure.

Evaluating and predicting therapy outcomes may be of great help to doctors in decision-making. An extensive revision of works about outcome prediction in patients with depression was carried out by Lee et al. [14]. Even more dramatic than mood disorders is the incidence of cancer in the population. Early detection is often believed to be the most effective treatment. Hence it is one of the significant targets of DL when applied to cancer therapy. In the case of colorectal cancer (CRC), the effectiveness of colonoscopy is measured by its adenoma detection rate, which can be increased by using convolutional neural networks (CNN) for image processing [15]. Furthermore, ML algorithms can predict the presence of lymph node metastasis, which represents a risk factor for patients who need surgery, thereby reducing the number of unnecessary operations[16]. Although many other studies applied ML and DL algorithms to medical contexts, being able to adapt a standard therapy to an individual patient, the so-called personalized therapies, remains a challenge [17, 18, 19, 20].

Although ML and DL models are powerful tools for decision-making in medicine and the number of clinical studies in the field of ML has grown tremendously in recent years [21, 22], they still lack the level of interpretation required to be understood by physicians. Thus, an extra layer should be added on top of the architecture to make the raw predictions provided by the models easier to understand for the general public. Unfortunately, this extra layer will imply an increase in computational needs and computing time. Hence, some parts of the workflow (data loading, training, prediction, and explanation) would have to be run on high-performance computing (HPC) platforms, and even then, the whole process might take several hours or days [6].

In this paper, we introduce a novel software tool called SIBILA, which aims to automate the training, evaluation and interpretation of many ML and DL models, all at once. To deal with performance issues, it can be run on HPC platforms, resulting in very competitive response times, which are a must in medical contexts. Unfortunately, not all HPC infrastructures are set up the same, which might result in problems running the software. To avoid these types of issues, SIBILA is containerized, so the HPC host only needs to support the container software to make SIBILA run. Moreover, our software applies a diversity of interpretability algorithms to provide the final user with a deep insight into the decisions that led the model to make a prediction. Therefore, with this tool, researchers may be able to save a great deal of time in looking for models to mine their datasets. Furthermore, they will be supplied with relevant information about the way predictions were made, which is a key factor for clinicians to be able to increase their knowledge. To make it more accessible, SIBILA features are also available through a freely accessible web server at https://bio-hpc.ucam.edu/sibila. Two case studies will be presented in this paper to discuss and prove the effectiveness of the tool.

## 2. Materials and methods

This section describes the main features of SIBILA, including the list of models and interpretability algorithms available, the way tasks are parallelized and how the consensus works. SIBILA has been programmed in Python3, and it relies on a set of widely used libraries to implement both the models and interpretability algorithms. In addition, all the packages are included within a Singularity container to make it portable.

SIBILA has been conceived to be modular and flexible so it can be extended in the future. Based on this idea, different modules have been implemented (Fig. 1). The blue boxes represent the modules the code is structured in. They are all surrounded by a continuous red line meaning that they are integrated into the core of SIBILA. In addition, some scripts are provided for pre- and post-processing tasks, such as the consensus. Note that the interpretability algorithms can be run in parallel if an HPC platform is available.

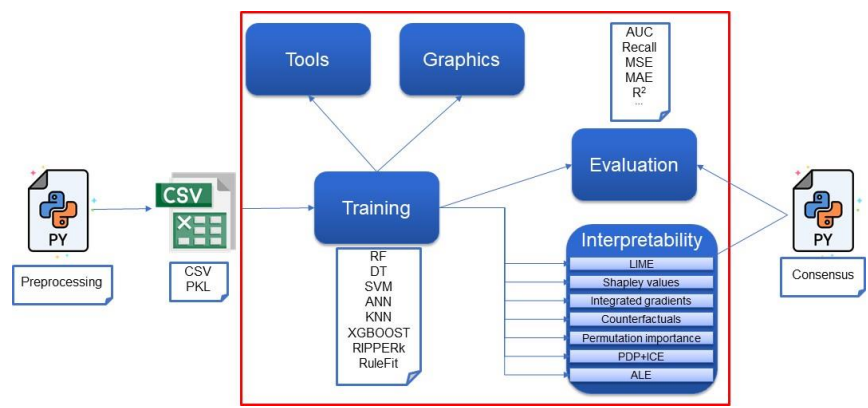

Figure 1: Schematic representation of the SIBILA modules.

### 2.1. Machine learning and deep learning models

SIBILA is an ensemble of ML and DL models that could easily be extended in the future. Selected models try to cover a wide range of approaches, including typical ML models (i.e. RF, DT), DL models (ANN) and rule-based models (RIPPER and RuleFit). Table 1 summarizes the available models, algorithms, libraries and whether they support regression and classification.

All models, including ANN, perform an automatic hyperparameter search that can be configured with JSON files.

Table 1: List of classification and regression models implemented by SIBILA.

| Model | Name | Library | Class. | Regress. | Ref. |
|---|---|---|---|---|---|
| DT | Decision Tree | scikit-learn | Yes | Yes | [23] |
| RF | Random Forest | scikit-learn | Yes | Yes | [23] |
| SVM | Support Vector Machine | scikit-learn | Yes | Yes | [23] |
| XGBOOST | eXtreme Gradient Boosting | xgboost | Yes | Yes | [24] |
| KNN | K-Nearest Neighbours | scikit-learn | Yes | Yes | [23] |
| ANN | Artificial Neural Network | TensorFlow 2, Keras Tuner | Yes | Yes | [25, 26] |
| RIPPER | Repeated Incremental Pruning to Produce Error Reduction | wittgenstein | Yes | No | [27] |
| RLF | RuleFit | rulefit | Yes | No | [28] |

## 2.2. Evaluation metrics

To assess the accuracy of the models, the typical metrics are calculated according to the nature of the problem (either classification or regression). Table 2 lists all available metrics.

Table 2: Evaluation metrics for classification and regression problems.

| Classification | Regression |
|---|---|
| Confusion matrix | Pearson coefficient |
| Accuracy | Coefficient of determination ($R^2$) |
| Precision | Mean absolute error (*MAE*) |
| F1 score | Mean squared error (*MSE*) |
| Recall | |
| Specificity | |
| Area Under the Curve (*AUC*) | |

## 2.3. Interpretability

As well as performance metrics, many interpretability algorithms are employed to detect the most relevant input features. The selected algorithms represent the attribution of each feature to the prediction numerically, either positive or negative. Additionally, all information regarding the attributions calculated by the algorithms will be bulked into CSV files for offline analysis. Table 3 shows these algorithms. It is not the aim of this manuscript to go into the algorithms, meaning only those details that are relevant for understanding the output of SIBILA will be provided.

Table 3: List of interpretability algorithms available in SIBILA.

| Approach | Library | Ref. |
| --- | --- | --- |
| Permutation importance | scikit-learn | [29] |
| Local Interpretable Model-Agnostic Explanations (LIME) | lime | [30] |
| Shapley values | shap | [31, 32] |
| Integrated gradients | alibi | [33] |
| Diverse Counterfactual Explanations (DiCE) | dice-ml | [34] |
| Partial Dependence Plot (PDP) + Individual Conditional Expectation (ICE) | scikit-learn | [35] |
| Accumulated Local Effects (ALE) | alibi | [36, 37] |

While PDP and ALE create individual graphs for each input variable, the rest of the methods only plot one, displaying the attribution of the inputs for every individual sample. Additionally, a global chart is created for local interpretation methods (i.e. LIME, Shapley values) to display the average attribution of the variables. This way users have an overview of the global attribution of the inputs.

*2.4. Containerization*

As mentioned above, SIBILA offers the possibility of being run either locally or in a high-performance computing environment. To facilitate its execution and increase its portability, a Singularity [38] container was created. As a result, it can be run on any computer that supports Singularity.

The choice of Singularity as the container instead of Docker [39] was based on two premises: i) the current trend of using Singularity instead of Docker in most HPC infrastructures [40]; ii) its availability on most of the HPC platforms we have access to.

*2.5. Consensus*

Although interpretability is strongly recommended in medical contexts, different algorithms may lead to contradictory explanations [41]. To reduce uncertainty in interpreting results, a consensus stage was implemented. Herein, the three types of consensuses provided by SIBILA are introduced. To do so, a synthetic dataset for binary classification was created to simulate a real input as proposed in a recent study [42]. The dataset, consisting of 17 input variables with uniformly distributed values in the range [0, 1] and 1948 samples, was built with one simple rule to determine the output class (Eq. 1). The aim of our tests was to identify such rule with the consensus functionality, thus $F3$ and $F6$ were expected to be returned as the most important features for the models. Note that consensus is an offline process that can be executed after interpretability is computed.

$$class = \begin{matrix} 1 \ if \ (0.7 \leq F3 \leq 0.9) \wedge (0.2 \leq F6 \leq 0.35) \\ 0 \ otherwise \end{matrix} \quad (1)$$

In the first type of consensus, the attributions of a single interpretability algorithm to the input features are averaged (Fig. 2a). Although it is a simple approach, it makes sense to get an overview of how each algorithm explains all the models. It is also believed that models with a low AUC may not be reliable, so a cut-off is introduced to disregard them and avoid incorrect interpretations. The default cut-off is 0.75, but the users can modify it. As shown in Figure 2a, according to Shapley values, $F6$ and $F3$ are the most relevant features because they have been given the highest attributions. This explanation matches the hidden rule of our synthetic dataset.

Although the attributions may be the best metric for assessing the importance of the variables, the interpretability algorithms applied may not be aligned, resulting in unequal attribution values. Based on this assumption, the second type of consensus averages the ranking of the input variables (Fig. 2b). Again, Figure 2b shows that $F6$ and $F3$ are clearly the best-ranked features and, consequently, the most important ones.

Finally, instead of grouping results by interpretability algorithm, all the explanations obtained for a given model are merged (Fig. 2c). Note that this consensus is based on the ranking of the features because each algorithm ranks attributions on a different scale. According to this calculation, $F6$ and $F3$ remain the most important features when using the ANN model.

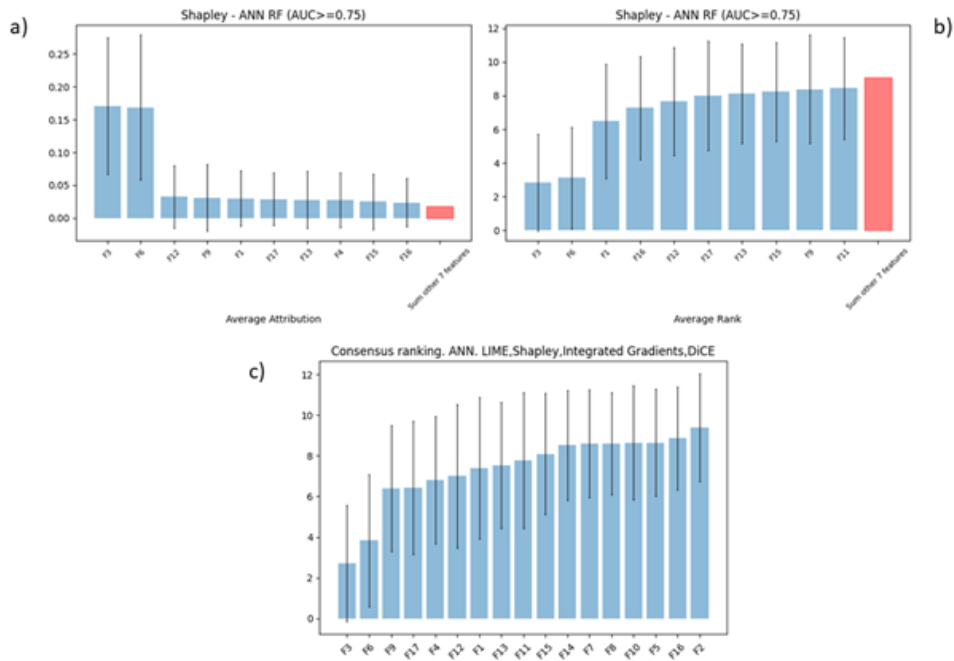

Figure 2: SIBILA calculates consensus in three different ways: a) By the average attribution of variables; b) By the relative ranking of the features; c) Combining all the samples of a model.

## 3. Results

This section introduces two case studies related to medicine. First, we will show the ability of SIBILA to predict the probability of a person of suffering from cancer. Then, we demonstrate that it can be useful in predicting the likelihood of suffering a heart attack. In addition, other two non-medical cases are supplied in the supplementary materials.

*3.1. Likelihood of cancer*

In this first case, the likelihood of a patient of suffering cancer was predicted. The original dataset, consisting of 1000 samples and 24 input features, was obtained from Kaggle [43]. The output class was a text field indicating the probability of suffering cancer, textually coded as Low, Medium and High probability.
Aiming to test SIBILA, which is a suitable tool for binary classification, the problem was transformed into a binary classification problem. As there was not enough information to include samples labeled Medium in the Low or High probability groups, they were removed from the dataset. Meanwhile, Low and High values were encoded as class 0 and 1 respectively. All other features were kept as numerical values, so no further processing was required. After cleaning, the final dataset was made of 303 (45.35%) patients with low and 365 (54.65%) with high probability of cancer.

Table 4: Classification metrics obtained from the models for the cancer dataset.

| Model | Accuracy | Precision | F1 | Recall | Specificity | AUC |
|---|---|---|---|---|---|---|
| ANN | 100±0.00 | 100±0.00 | 100±0.00 | 100±0.00 | 100±0.00 | **1±0.00** |
| DT | 100±0.00 | 100±0.00 | 100±0.00 | 100±0.00 | 100±0.00 | **1±0.00** |
| SVM | 100±0.00 | 100±0.00 | 100±0.00 | 100±0.00 | 100±0.00 | **1±0.00** |
| RF | 100±0.00 | 100±0.00 | 100±0.00 | 100±0.00 | 100±0.00 | **1±0.00** |
| XGBOOST | 100±0.00 | 100±0.00 | 100±0.00 | 100±0.00 | 100±0.00 | **1±0.00** |
| KNN | 100±0.00 | 100±0.00 | 100±0.00 | 100±0.00 | 100±0.00 | **1±0.00** |
| RP | 47.04±3.74 | 100±0.00 | 1.40±0.10 | 2.75±0.19 | 100±0.00 | 0,51±0.00 |
| RLF | 100±0.00 | 100±0.00 | 100±0.00 | 100±0.00 | 100±0.00 | **1±0.00** |

Each model was trained 50 times with random seed to avoid biased results. The only difference in the configuration was that, while ANN was trained with 10 folds of cross-validation, the other models required only 5 folds to achieve the same accuracy. Once all models were trained, the classification metrics calculated by SIBILA were averaged and their standard deviations computed (Table 4). Next, the interpretability of all the selected models was calculated. The full list with the top 10 relevant features for each model are available as supplementary material (Tables S3-S10).

As can be seen in Table 4, all models except RIPPER achieved a perfect classification rate. This is due to the simplicity of the data. Despite of this, RIPPER was hardly able to improve the accuracy of a random classifier. Consequently, it was not considered for interpretability purposes.

Finally, consensus is computed for all the selected models (Fig. 3). It can be noticed that the set of relevant features is similar in all the models, what was the expected result of the first consensus. Going a step further, *Gender*,

*Age* and *Air pollution* were identified as important in all models, while *Dust allergy* and *Alcohol use* in six out of the seven models. In contrast, features that might be expected to be relevant, such as *Genetic risk*, *Smoking* or *Fatigue* do not to be as important as one might think.

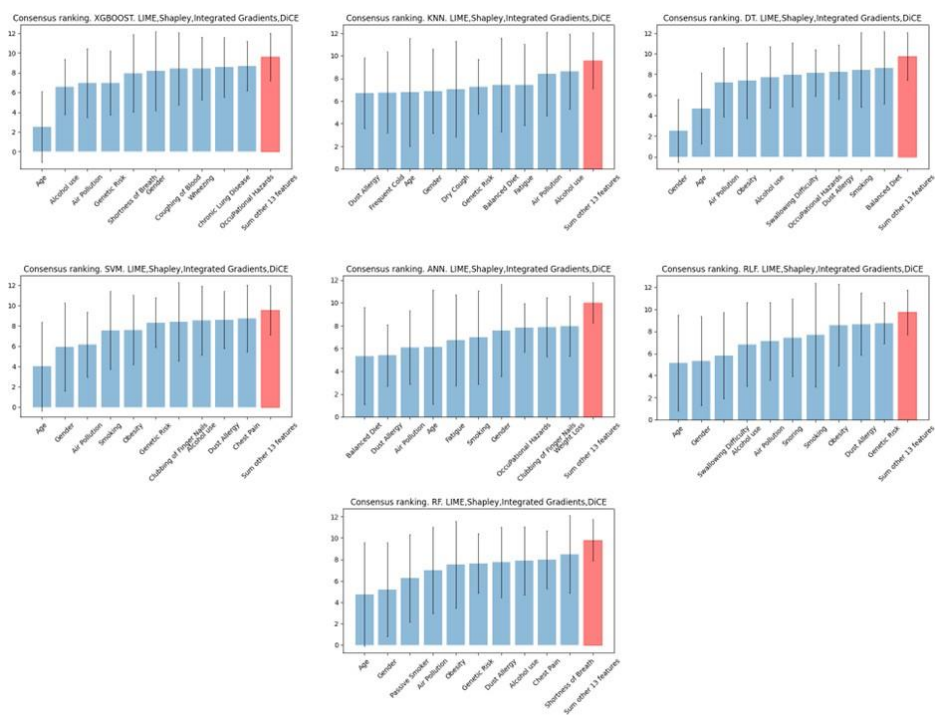

Figure 3: Top-10 most relevant features for each model trained on the cancer dataset.

## 3.2. Heart attack prediction

In the following example, we tried to predict the probability of a patient suffering a heart attack. Data was imported from the Kaggle repository [44] and curated to model it as a binary classification problem. All the models were trained 30 times and the metrics were averaged.

After training all models, the highest-scoring ones obtained similar accuracy: RF (*AUC* = 0.818), RLF (*AUC* = 0.806), ANN, RP and XGBOOST (*AUC* = 0.795). In contrast, SVM (*AUC* = 0.537) achieved the worst result (Table 5). Aiming to avoid bad models being part of the explanation, only the models with an accuracy higher than a certain threshold (*AUC* $\geq$ 0.75) were considered for the consensus process. Table 6 shows the most relevant features obtained for each model after having applied consensus.

| Model | Accuracy | Precision | F1 | Recall | Specificity | AUC |
|---|---|---|---|---|---|---|
| ANN | 79.348±0.00 | 82.979±0.00 | 80.412±0.00 | 78.000±0.00 | 80.952±0.00 | **0.795±0.00** |
| DT | 77.174±0.00 | 76.852±0.00 | 79.808±0.00 | 83.000±0.00 | 70.238±0.0001 | 0.766±0.00 |
| SVM | 57.609±0.00 | 56.250±0.00 | 71.739±0.00 | 99.000±0.00 | 8.333±0.00 | 0.537±0.00 |
| RF | 85.537±1.09 | 80.297±0.89 | 84.843±0.94 | 89.933±0.99 | 73.730±1.19 | **0.818±0.01** |
| XGBOOST | 80.978±0.00 | 75.194±0.00 | 84.716±0.00 | 97.000±0.00 | 61.905±0.00 | **0.795±0.00** |
| KNN | 72.283±0.00 | 72.477±0.00 | 75.598±0.00 | 79.000±0.00 | 64.286±0.00 | 0.716±0.00 |
| RP | 80.435±0.00 | 77.586±0.00 | 83.333±0.00 | 90.000±0.00 | 69.048±0.00 | **0.795±0.00** |
| RLF | 81.286±1.78 | 79.365±1.22 | 83.716±1.67 | 88.600±2.68 | 72.580±1.73 | **0.806±0.02** |

Table 5: Classification metrics obtained from the models for the heart attack dataset.

Table 6: Top-10 attributed features identified by each interpretability algorithm.

| LIME | Shapley | Integrated Gradients | DiCE |
|---|---|---|---|
| ST Slope | ST Slope | MaxHR | ST Slope |
| ChestPainType ASY | ChestPainType ASY | Cholesterol | Cholesterol |
| Cholesterol | Cholesterol | RestingBP | ChestPainType ASY |
| Oldpeak | Oldpeak | Age | Oldpeak |
| ExerciseAngina | MaxHR | | MaxHR |
| Age | Age | | ExerciseAngina |
| MaxHR | ExerciseAngina | | Age |
| RestingBP | RestingBP | | RestingBP |
| RestingECG Normal | RestingECG Normal | | RestingECG Normal |
| Sex | Sex | | Sex |

We observed that all the methods focused on the same set of variables. To confirm this finding, we analyzed the average attributions given by any of the algorithms to RF, which was the best-ranked model (Fig. 4). To prove the validity of the different algorithms, we chose LIME for this test.

In order to contrast the explanation given by LIME and obtain the edge values of each feature, its explanation was compared with a rule-based model. In this case, the equations obtained from RIPPER were examined and summarized in one single equation (Eq. 2). As can be seen, most of the features involved in Eq. 2 are also present in Figure 4 what confirms that the explanations were correct.

$$\begin{aligned}
\text{class} = 1 \text{ if } & [\text{ChestPainType = Asymptomatic} \wedge \text{ST\_Slope = Upsloping} \\
& \quad \wedge \text{Cholesterol} \leqslant 149.0] \\
& \vee [\text{ST\_Slope = Upsloping} \wedge \text{ExerciseAngina}] \\
& \vee [\text{ST\_Slope = Upsloping} \wedge \text{ChestPainType = Asymptomatic} \\
& \quad \wedge \text{Oldpeak} \leqslant 0.0] \\
& \vee [\text{ST\_Slope = Upsloping}] \\
& \vee [\text{ChestPainType = Asymptomatic} \wedge \text{ST\_Slope = Flat}] \\
& \vee [\text{ChestPainType = Asymptomatic} \wedge \text{Oldpeak} \in [1.5, 1.9]] \\
& \vee [\text{ChestPainType = Asymptomatic} \wedge \text{ExerciseAngina} \\
& \quad \wedge \text{RestingBP} \in [120, 130]] \\
& \vee [\text{Cholesterol} \leqslant 149.0 \wedge \text{ChestPainType = Asymptomatic} \\
& \quad \wedge \text{RestingECG} \quad \text{Normal}] \\
& \vee [\text{Cholesterol} \leqslant 149.0 \wedge \text{RestingBP} \leqslant 110.0] \\
& \vee [\text{Cholesterol} \leqslant 149.0 \wedge \text{Oldpeak} \geqslant 2.37] \\
& \vee [\text{Cholesterol} \leqslant 149.0 \wedge \text{ExerciseAngina}] \\
& \vee [\text{Age} \in [57, 59] \wedge \text{RestingBP} \in [130, 136]] \\
& \vee [\text{Oldpeak} \in [1.0, 1.5] \wedge \text{Age} \geqslant 65] \\
0 \text{ otherwise} &
\end{aligned} \quad (2)$$

## 4. Discussion

To demonstrate the usefulness of SIBILA, two case studies were presented. In the first case, the probability of cancer was predicted and the most relevant features were identified. As might be expected, each interpretability algorithm put the eye on different features when explaining the same model. This is the so-called disagreement problem, which prevents users from understanding the inner workings of the models. To overcome this problem, a consensus process was applied. As a a result, three features (*gender*, *age* and *air pollution*) were found to be present in the explanation of all models. Therefore, they could be asserted as frequent indicators for determining the likelihood of developing cancer. The novelty of this sample is that a clinician could have identified such features just by quickly having a look at the consensus charts.

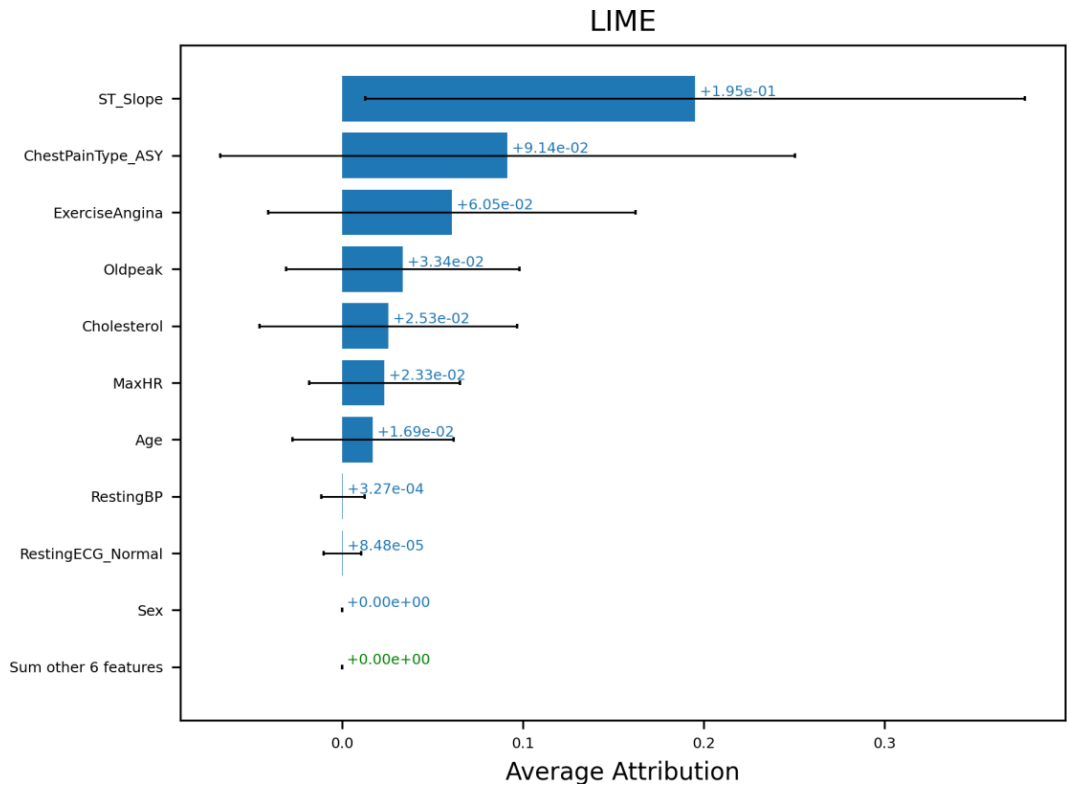

Figure 4: Top 10 features attributed by LIME to the RF model. The X-axis shows the averaged attribution of each feature, and the Y-axis represents the top 10 most important features. Black horizontal lines depict the standard deviation of the attribution values given to the features.

In the last case, a dataset with samples of patients who either did or did not suffer a heart attack was used to simulate a real case. Five out of eight models outperformed the other models with an AUC higher than 0.795. Among the most accurate models, RF performed the best ($AUC$ = 0.818). We therefore used it as a baseline to explore the results. The consensus identified a subset of features that was able to explain any test sample of the dataset. After that, the average explanation given by LIME to the RF model was compared with that particular subset and confirmed the validity of the variables. In addition, the RIPPER algorithm was used to determine the edge values of the features and summarize all the conditions into one single equation. As shown in Eq. 2, *high blood pressure* and *exercise-induced angina* are risk factors for suffering a heart attack. Also, not suffering from *chest pain* might be dangerous when combined with other factors such as *high blood pressure* and *oldpeak*. It is worth mentioning that RIPPER did not pay any attention to the *maximum heart rate* factor, although it might be expected to be a risk factor.

Although the features identified require further validation by experts, it should be highlighted that the use of consensus helped us find a consistent set of features explaining how the models worked internally. In addition, even though previous studies on the heart attack dataset [45, 46] reported a similar set of relevant features, our tool was able to depict the models' decision process in a very understandable way.

## 5. Conclusions

ML and DL are powerful tools for decision-making systems, but their lack of interpretability remains a significant challenge. Without interpretability, physicians, lawyers, brokers and politicians, among others, will never trust their predictions, and society will miss the opportunity to take advantage of these techniques. With the aim of bringing this scenario to light, an ensemble of ML/DL models for training datasets and explaining predictions has been developed. SIBILA saves scientists a lot of time in building ad-hoc models for each dataset. Indeed, it automatically computes interpretability, meaning the results are displayed in an understandable way for non-IT users, such as healthcare staff. Bearing these types of users in mind, the features of SIBILA have been delivered at no cost at: https://bio-hpc.ucam.edu/sibila.

The results prove that SIBILA is able to quickly train and explain models on both classification and regression problems. In addition, using the consensus after interpretability is calculated, may be a powerful tool for the identification of the most relevant features. Even when the interpretability algorithms disagree, the consensus calculates a global explanation that covers all the individual samples. This overall explanation is very important for non-IT users to be able to understand what is happening inside the models. Furthermore, the training of large datasets may take several hours or days, which is a drawback in critical contexts, such as medicine. To cope with this inconvenience, the entire pipeline can be computed on separate jobs of an HPC cluster. This requires no extra configuration because the code of SIBILA is containerized.

Despite its scope of application not being restricted to medicine and the fact that it can be used with any dataset, SIBILA is thought to help doctors in decision-making, which is why it was tested with medical datasets. Consequently, we truly believe that it could be helpful in the treatment of diseases such as diabetes or COVID-19. More importantly, it could be applied to biological or pharmaceutical datasets to find unknown biomarkers and effective drug combinations.

**Source code**

The source code is available at https://github.com/bio-hpc/sibila.git.

**Funding**

This work has been funded by grants from the European Project Horizon 2020 SC1-BHC-02-2019 [REVERT, ID:848098]; Fundación Séneca del Centro de Coordinación de la Investigación de la Región de Murcia [Project 20988/PI/18]; and the Spanish Ministry of Economy and Competitiveness [CTQ2017-87974-R].

**Credit authorship contribution statement**

**Antonio Jesús Banegas-Luna**: Conceptualization, Investigation, Data curation, Software, Writing. **Horacio Pérez-Sánchez**: Funding, Conceptualization, Investigation, Validation, Supervision, Writing.

**Declaration of Competing Interest**

The authors declare no competing interest.

**Acknowledgements**

This work has been funded by grants from the European Project Horizon 2020 SC1-BHC-02-2019 [REVERT, ID:848098]; Fundación Séneca del Centro de Coordinación de la Investigación de la Región de Murcia [Project 20988/PI/18]]; and the Spanish Ministry of Economy and Competitiveness [CTQ2017-87974-R]. Supercomputing resources in this work were supported by the Poznan Supercomputing Center's infrastructures, the e-infrastructure program of the Research Council of Norway, and the supercomputing centre of UiT—the Arctic University of Norway, the Plataforma Andaluza de Bioinformática at the University of Málaga, the supercomputing infrastructure of the NLHPC (ECM-02, Powered@NLHPC), and the Extremadura Research Centre for Advanced Technologies (CETA-CIEMAT), funded by the European Regional Development Fund (ERDF). CETA-CIEMAT is part of CIEMAT and the Government of Spain.